\pdfoutput=1

\documentclass[11pt]{article}

\usepackage[preprint]{acl}
\usepackage[ruled,vlined]{algorithm2e}
\usepackage{amsmath,amssymb,amsfonts}
\usepackage[noend]{algpseudocode}
\usepackage{pifont}
\usepackage{wrapfig}
\usepackage{bm}	
\usepackage{bbm}
\usepackage{float}
\usepackage{newfloat}
\usepackage{array}
\usepackage{placeins}
\usepackage{listings}
\usepackage{nicefrac} 
\usepackage{graphicx}
\usepackage{bbding}
\usepackage{dsfont}
\usepackage{mathtools}
\usepackage{fontawesome}
\usepackage{makecell}
\usepackage{multirow}
\usepackage{ragged2e}
\usepackage{color, colortbl}
\usepackage{xcolor}
\usepackage{subcaption}
\usepackage{microtype}
\usepackage{hyperref}
\usepackage{url}
\usepackage{booktabs}

\usepackage{times}
\usepackage{latexsym}

\usepackage[T1]{fontenc}

\usepackage[utf8]{inputenc}

\usepackage{microtype}

\usepackage{inconsolata}

\usepackage{graphicx}
\newcommand{\gr}{\rowcolor[gray]{.95}}
\usepackage{arydshln}
\newcommand{\W}{\boldsymbol{W}}
\newcommand{\m}{\boldsymbol{M}}

\title{
Rethinking Pruning for Vision-Language Models: Strategies for Effective Sparsity and Performance Restoration
}

\author{
    Shwai He\textsuperscript{\rm 1} \space\space Ang Li\textsuperscript{\rm 1} \space\space Tianlong Chen\textsuperscript{\rm 2}\thanks{~Corresponding author} \\
    \textsuperscript{\rm 1}University of Maryland, College Park \\
    \textsuperscript{\rm 2}University of North Carolina at Chapel Hill \\
    {\tt\small shwaihe@umd.edu}, \space
    {\tt\small angliece@umd.edu}, \space
    {\tt\small tianlong@cs.unc.edu}
        }

\begin{document}
\maketitle
\begin{abstract}
Vision-Language Models (VLMs) integrate information from multiple modalities and have shown remarkable success across various tasks. However, deploying large-scale VLMs in resource-constrained scenarios is challenging. Pruning followed by finetuning offers a potential solution but remains underexplored for VLMs. This study addresses two key questions: how to distribute sparsity across different modality-specific models, and how to restore the performance of pruned sparse VLMs.
Our preliminary studies identified two effective pruning settings: applying the same sparsity to both vision and language models, and pruning only the language models. 
While LoRA finetuning aims to restore sparse models, it faces challenges due to incompatibility with sparse models, disrupting the pruned sparsity.
To overcome these issues, we propose SparseLoRA, which applies sparsity directly to LoRA weights. Our experimental results demonstrate significant improvements, including an 11.3\% boost under 2:4 sparsity and a 47.6\% enhancement under unstructured 70\% sparsity. 
Code is released at: \url{https://github.com/Shwai-He/VLM-Compression}. 
\end{abstract}

\section{Introduction}

Scaling deep learning models has demonstrated promising performance across various tasks in both vision and language domains \cite{brown2020language, mixtral, zhu2023minigpt4}. Vision-Language Models (VLMs) \cite{clip, blip, liu2023visual}, which leverage powerful vision and language models, have recently garnered significant attention in research \cite{gpt4, liu2024sora}, showcasing their cross-modality capabilities. However, the ever-increasing size of these models comes with substantial computational and memory costs, limiting their practical applicability in resource-constrained environments. Model pruning followed by finetuning \cite{dai2018vib, fang2023depgraph, tanaka2020pruning}, which reduces model size while preserving performance, holds promise for improving the real-world deployment of VLMs.

\begin{figure*}
\centering
\vspace{-5pt}
\hspace{-4pt}\begin{minipage}[b]{0.48\linewidth}
\includegraphics[width=\linewidth]{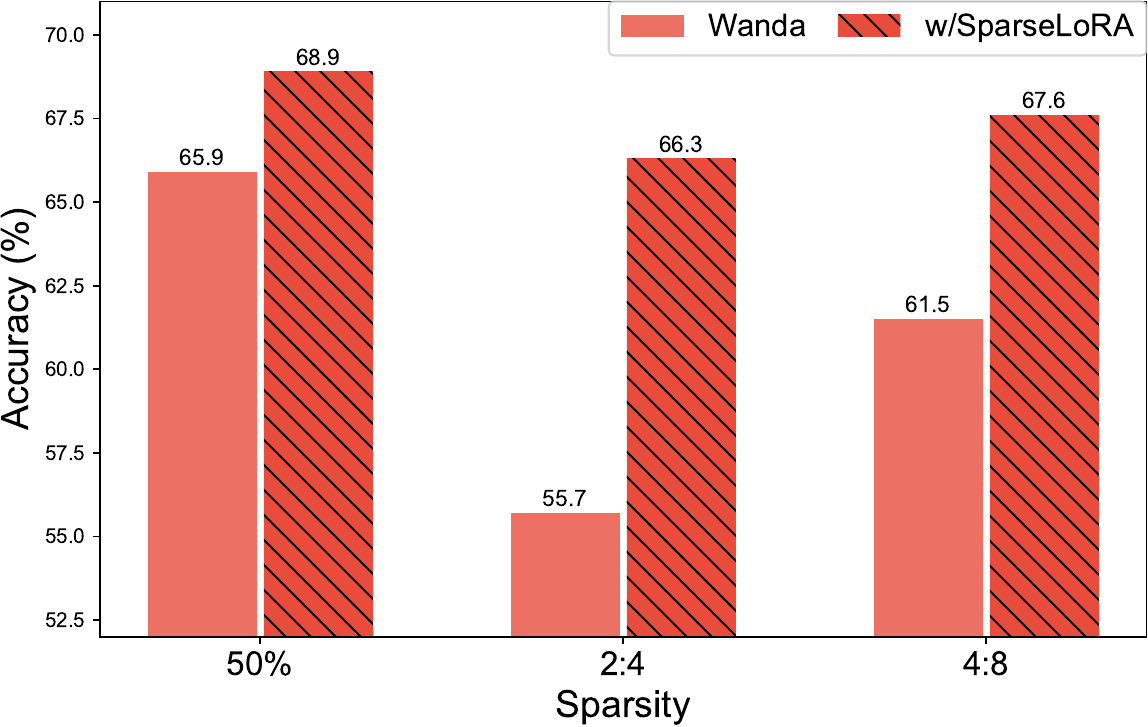} 
\subcaption{InstructBLIP-$\text{FlanT5}_\text{XL}$}
\end{minipage}
\begin{minipage}[b]{0.48\linewidth}
\includegraphics[width=\linewidth]{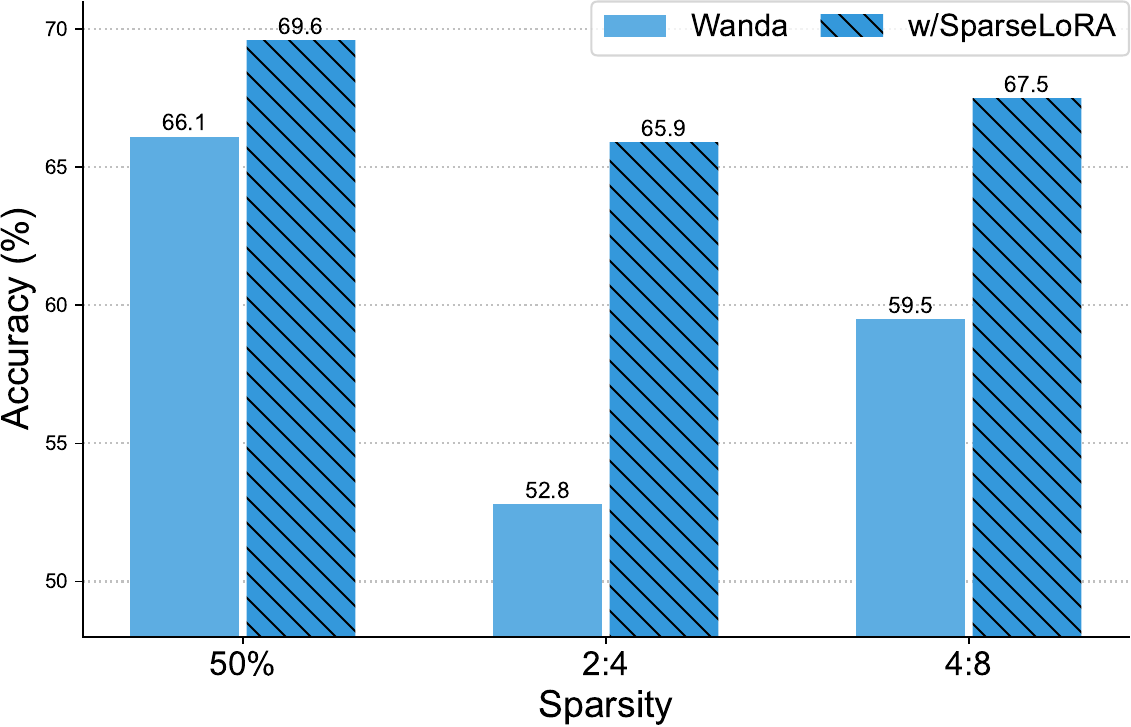}
\subcaption{InstructBLIP-Vicuna-7B}
\end{minipage}
\caption{\textbf{The comparison of pruned VLMs (``Wanda'') and restored VLMs (``w/SparseLoRA'')} on multimodal tasks, taking InnstructBLIP \cite{instructblip} as the backbone.  }
 \vspace{-5pt}
\label{fig:highlights}
\end{figure*}

While pruning followed by finetuning has significantly improved the efficiency of vision models \cite{frankle2019lottery, kusupati2020soft, lee2018snip} and language models \cite{chen2020lottery, wanda, sparsegpt}, the realm of Vision-Language Models (VLMs) remains relatively unexplored in terms of model pruning, prompting following questions: \textit{how to distribute sparsity ratios between different modality-specific models} and \textit{how to restore the performance of prune sparse VLMs}. 

For the first question, we conducted empirical studies on pruning modality-specific models, experimenting with various combinations of sparsity ratios. Surprisingly, we found that applying the same sparsity ratios to both vision and language models yields nearly optimal performance. On the other hand, since language models are usually much larger than vision models, pruning only the language models offers a beneficial trade-off between performance and efficiency. However, as sparsity ratios increase, pruning significantly degrades performance, especially with structured sparsity patterns (e.g., N: M sparsity \cite{zhang2022learning, zhou2021}), underscoring the importance of post-pruning restoration.

While parameter-efficient LoRA finetuning has been proposed to repair the performance of sparse models, it faces a significant challenge due to the incompatibility of dense LoRA modules with sparse models. Merging LoRA modules with sparse models would destroy the sparse pattern, while maintaining LoRA modules would introduce extra latency and slow down the inference speed.
To address the incompatibility issue of LoRA, we introduce SparseLoRA finetuning, which utilizes binary masks on LoRA weights, allowing seamless integration with pruned weights. 

Extensive experiments showcase the effectiveness of our proposed methods in repairing the performance of pruned sparse VLMs. For instance, as illustrated in Figure \ref{fig:highlights}, SparseLoRA boosts the performance by 13.1\% for InstructBLIP-Vicuana-7B with 2:4 sparsity. In summary, our contributions are threefold:

\begin{itemize}
\item We empirically study the modality-specific sparsity distributions and systematically demonstrate how sparsity affects the performance of VLMs.
\item We propose a pipeline involving pruning and post-finetuning with SparseLoRA to restore pruned models.
\item Extensive experiments validate the effectiveness and universality of SparseLoRA across various VLMs and tasks.
\end{itemize}

\section{Related Work}

\textbf{Vision-Language Models. }
Vision-language models, among the most sophisticated multi-modal architectures, have demonstrated outstanding performance across various cross-modality tasks, including image captions \cite{cc3m}, image retrieval \cite{Plummer2015Flickr30kEC}, visual QA \cite{Kim2016MultimodalRL}, and image/video generation \cite{zhou2021lafite, singer2022makeavideo}. 
These models typically freeze the pretrained vision and language components, only fine-tuning a small, learnable interface (e.g., Qformer in BLIP-2 \cite{blip-2}) to facilitate inter-modality interactions \cite{yin2023survey, blip-2}, thus avoiding high training costs and potential catastrophic forgetting \cite{forgetting}. 

\textbf{Model Pruning for Large Language Models. }
While large vision and language models have shown promising advancements, their massive parameter sizes present challenges for practical deployment \cite{llm-pruner, wang-etal-2020-structured}. To mitigate this, model pruning techniques have been introduced to remove redundant weights or structures \cite{han2016deep, NIPS2016_6e7d2da6}. The primary aim of model pruning is to minimize the disparity between models before and after pruning \cite{liu2021group, He_2024, sparsegpt}. Various metrics, such as magnitude, gradient \cite{ECoFLaP}, and activation \cite{wanda}, have been proposed to identify unimportant weights. However, pruning without finetuning often leads to a performance drop. \cite{dsnot} utilize reconstruction errors-based metrics to update the weights. Other than the disparity between sparse models and dense models, our method also considers the task-specific objective of repairing sparse models and knowledge distillation from the original full models.

\section{Preliminary Study}
\begin{figure*}[h]
\centering
\hspace{-8pt}
\begin{minipage}[b]{0.46\linewidth}
\includegraphics[width=\linewidth]{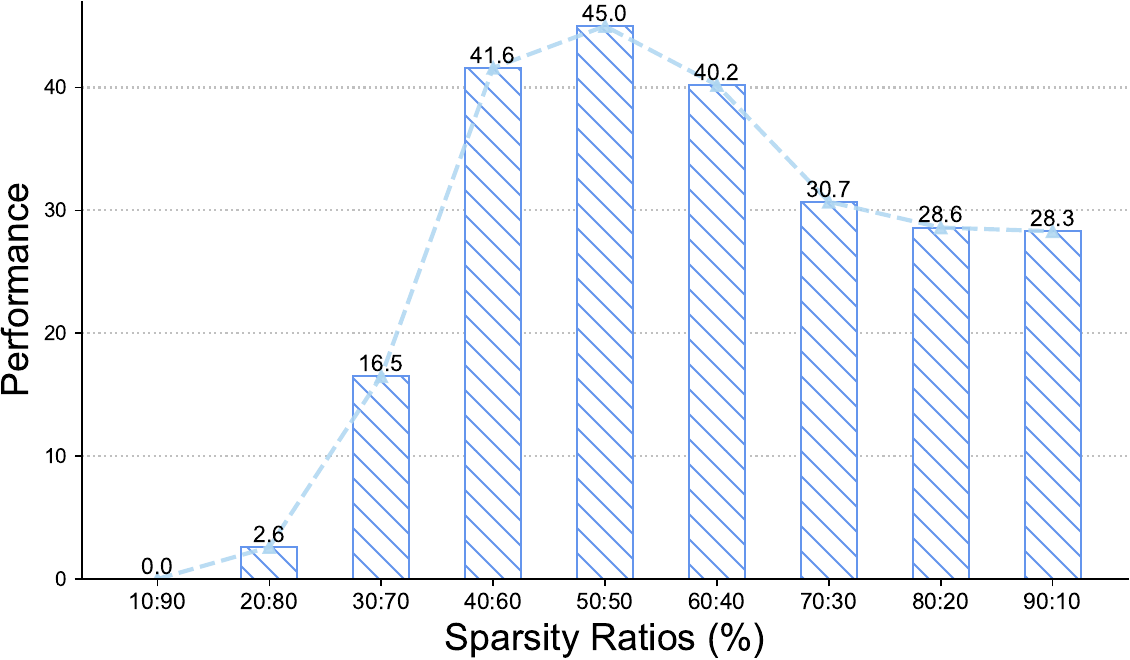} 
\subcaption{$s_v + s_l = 100\%$}
    \end{minipage}\hspace{8pt}\begin{minipage}[b]{0.46\linewidth}
\includegraphics[width=\linewidth]{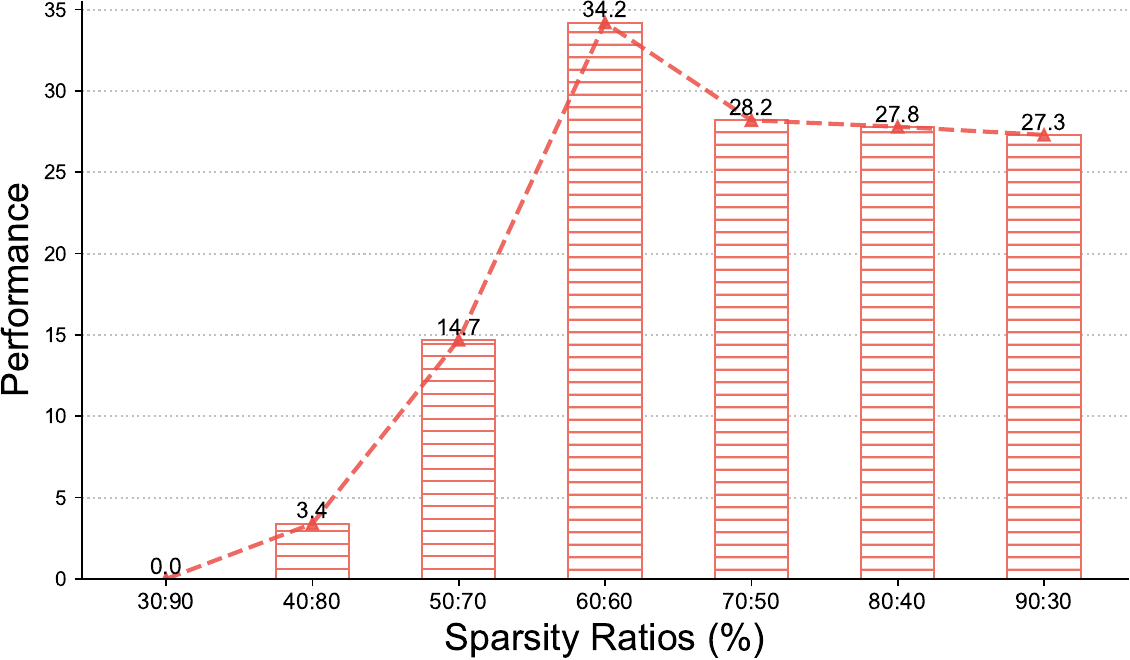}
\subcaption{$s_v + s_l = 120\%$}
        \end{minipage}
\caption{\textbf{Performance of BLIP-2 with different modality-specific sparsity distribution}. We denote the sparsity ratios for the vision and language modalities as ``$s_v$:$s_l$". We adjust their distribution while constraining their summation ``$s_v + s_l$'' to be (a) $100\%$ and (b) $120\%$. }
\label{fig:sparsity_combination}
\end{figure*}
\begin{figure*}[h]
\centering
\begin{minipage}[b]{0.46\linewidth}
\hspace{-5pt}\includegraphics[width=\linewidth]{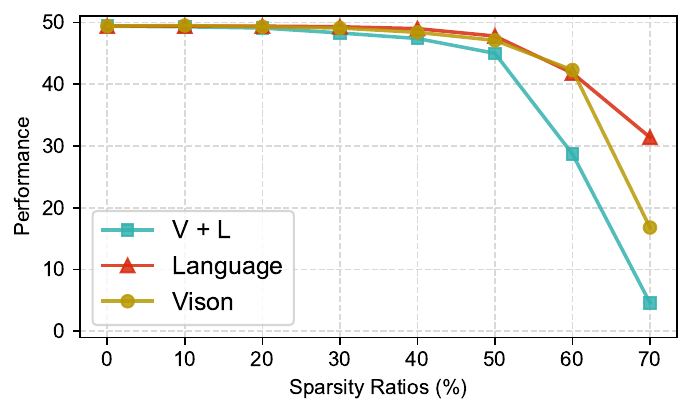} 
\subcaption{Unstructured Sparsity}
\label{fig:unstructured_sparsity}
\end{minipage}
\begin{minipage}[b]{0.46\linewidth}
\includegraphics[width=\linewidth]{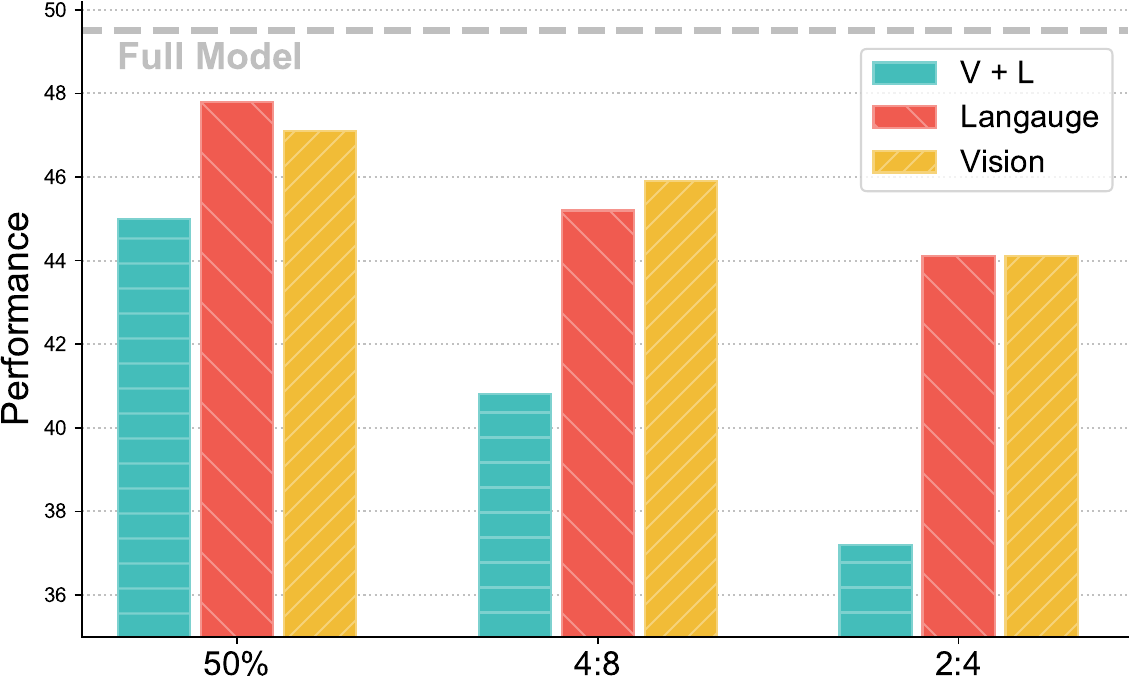}
\subcaption{Structured Sparsity}
\label{fig:structured_sparsity}
        \end{minipage}
\vspace{-8pt}
\caption{\textbf{Performance of BLIP-2 with different sparse ratios} (i.e., unstructured pruning and N:M pruning) for visual question-answering tasks. }
\vspace{-4pt}
\end{figure*}

Vision-Language Models (VLMs) consist of modality-specific foundation models, namely visual and language models, as well as a cross-modality interface (e.g., QFormer \cite{blip-2}) that aligns models from different modalities. Following \cite{ECoFLaP}, we focus on pruning the vision and language models while keeping the Q-Former intact, as it is sufficiently lightweight. 
Parameters are not evenly distributed across the different modality-specific models; for instance, visual models are often considerably smaller than their corresponding language models \cite{blip-2, instructblip, liu2023visual, yang2021empirical}. In this case, we pose two questions: (1) {\textit{how to distribute sparsity ratios between modality-specific models}, and (2) \textit{how do different sparsity ratios affect the performance of VLMs?}}

For the first question, we first try various sparsity ratio combinations between visual models and language models. Specifically, we fix the summation of $s_v$ and $s_l$ and then adjust their distributions accordingly. We use Wanda \cite{wanda} as the default pruning method because it ensures relatively high performance and efficiency. Based on Figure \ref{fig:sparsity_combination}, we found: 
(1) VLMs would collapse when the language models are under high sparsity ratios (i.e., $s_l > 70\%$), whereas sparsity imposed on visual models has a comparatively lower impact on performance; 
(2) When constrained by the summation of sparsity ($s_v + s_l$), pruning the modality-specific models with equal sparsity ratios leads to optimal performance.

For the second question, we initially prune VLMs with different unstructured sparsity ratios using the following strategies: pruning language models and visual models with the same sparsity ratios (``V + L''), pruning visual models only (``Vision''), and pruning language models only (``Language''). According to Figure \ref{fig:unstructured_sparsity}, when sparsity ratios exceed 50\%, all settings experience a significant performance drop, although VLMs pruned by a single modality model maintain relatively high performance.

Similarly, in Figure \ref{fig:structured_sparsity} when employing structured N:M sparsity \cite{zhou2021, zhang2022learning} (i.e., in each contiguous block of $M$ values, $N$ values must be zero), all models encounter significant performance degradation and even collapse (2:4 for pruning both vision models and language models). 
This situation prompts us to reflect on \textit{how to restore the pruning-caused performance degradation for VLMs}.

\begin{figure*}[h]
\centering
\includegraphics[width=0.96\linewidth]{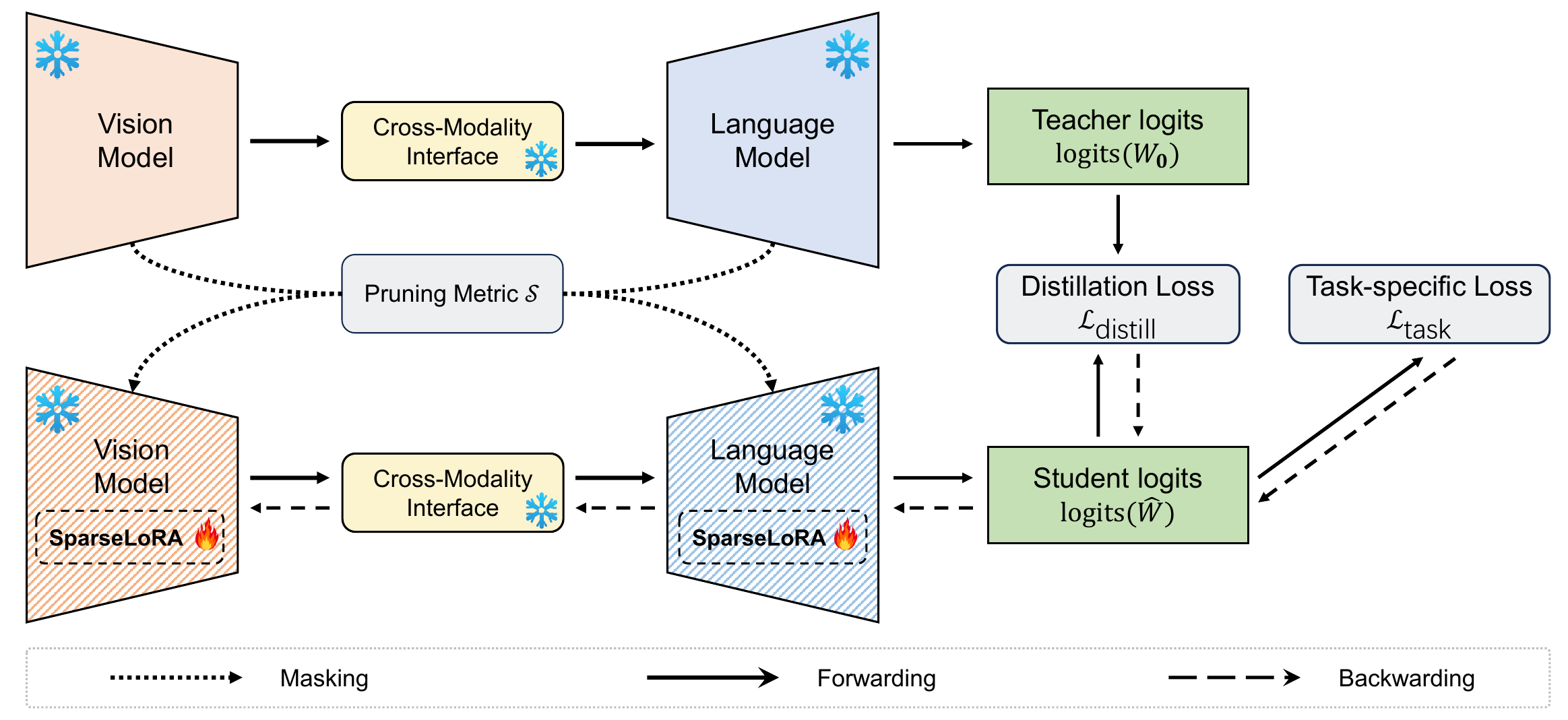}
\vspace{-5pt}
\caption{\textbf{Visualization of the pipeline of VLM Pruning and SparseLoRA finetuning,} which first prunes the vision model and language model based on a given pruning metric, then restores the pruned via SparseLoRA finetuning. }
\label{fig:overview}
\end{figure*}

\section{Methodology}

In this section, we will develop a pipeline that involves pruning and post restoration, with the illustration in Figure \ref{fig:overview}.

\subsection{Pruning with Few Samples}
Model pruning identifies less important weights using predefined metrics \cite{han2016deep}, typically measuring the reconstruction errors \cite{wanda, sparsegpt, dsnot} between models before and after pruning, such as magnitude, gradient, and activation. Calculating gradients or activation requires a small calibration dataset $\mathcal{D}_p$ with few samples. With predefined metric $\mathcal{S}$ and the calibration dataset, the weights of a model is scored as follows:
\begin{equation}
    S \leftarrow \mathcal{S}(\W_0, \mathcal{D}_p), 
    \label{eq:score}
\end{equation}
where $\W_0$ denote the weights of the model while $S$ represent the importance scores for $\W_0$. Given the sparse ratios $s$, binary masks are utilized to locate the pruned weights and update the weights as follows: 
\begin{equation}
    \m \leftarrow (S > \tau ), \quad \W \leftarrow \W_0 \odot \m,
    \label{eq:mask}
\end{equation}
where $\W$ denotes the pruned weights, while \textbf{$\tau$} represents the threshold ($s$ percentile of ${S}$) and all weights with scores lower than $s$ will be removed. While the pruning metrics $\mathcal{S}$ aim to minimize reconstruction errors \cite{wanda, dsnot} or maintain performance \cite{ECoFLaP}, model pruning often results in a significant performance drop and therefore needs to be recovered.

\subsection{Sparse LoRA finetuning}

\begin{figure*}[h]
\centering
\hspace{-4pt}\begin{minipage}[b]{0.46\linewidth}
\includegraphics[width=\linewidth]{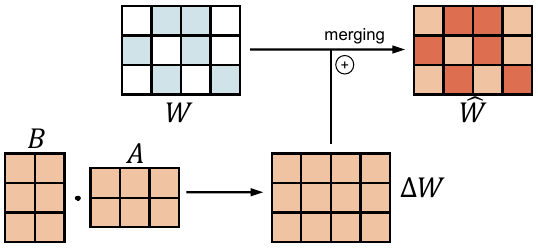} 
\subcaption{LoRA}
    \end{minipage}\hspace{18pt}\begin{minipage}[b]{0.46\linewidth}
\includegraphics[width=\linewidth]{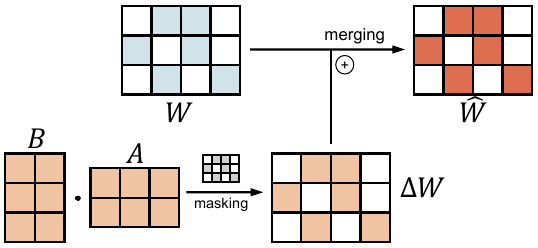}
\subcaption{SparseLoRA}
        \end{minipage}
\caption{\textbf{Schematic comparison of (a) LoRA and (b) SparseLoRA.} With masking, SparseLoRA preserves the sparse patterns, while LoRA destroys them after merging.}
 \vspace{-10pt}
\label{fig:SparseLoRA}
\end{figure*}

VLMs, which incorporate both vision models and language models, are often too large to be fine-tuned through full-model fine-tuning techniques \cite{blip-2, instructblip}. Instead, parameter-efficient fine-tuning techniques \cite{houlsby2019parameterefficient, peft, lora} are employed to reduce the number of trainable parameters while maintaining comparable performance. Among these techniques, LoRA \cite{lora} stands out as one of the most widely used approaches. since it not only efficiently utilizes parameters but also allows for seamless integration with the original weights, thus avoiding potential latency during inference \cite{dery2024everybody, rücklé2021adapterdrop}.

Traditional LoRA fine-tuning involves freezing the parameters of the pretrained model and injecting trainable rank decomposition matrices into each layer that requires fine-tuning. LoRA modules involves two small low-rank trainable weights $A$ and $B$, which can be merged with $\W$ after finetuning: 
\begin{equation}
    \W \leftarrow \W + \Delta \W, \quad \text{where  } \Delta \W = BA.  
\end{equation}
However, as shown in Figure \ref{fig:SparseLoRA}, the sparse pattern of pruned models would collapse after merging \cite{dery2024everybody, he2023sdconv}. Given that $\Delta \W$ are dense weights and $\W$ are sparse weights, the element-wise operation would destroy the sparse patterns. Additionally, without merging, the injected LoRA modules would increase latency and slow down inference speed \cite{mundra2023comprehensive, rücklé2021adapterdrop, dery2024everybody}. Inspired by \cite{sparseadapter}, we propose employing masks on $\W$ to preserve the sparse pattern:  
\begin{equation}
    \hat{\boldsymbol{W}} = \boldsymbol{W} + \Delta \W \odot \m. 
\end{equation}
In such a case, $\Delta \W$ corresponding to pruned positions are masked and cannot be updated via gradient-backpropagation. Consequently, during backpropagation, during backpropagation, $A$ and $B$ can be optimized as follows: 
\begin{equation}
\begin{aligned}
    B \leftarrow B + \eta \cdot (\frac{\partial \mathcal{L}}{\partial \hat{\W}} \odot \m)A^T, \\
    A \leftarrow A + \eta \cdot B^T (\frac{\partial \mathcal{L}}{\partial \hat{\W}} \odot \m), 
\end{aligned}
\end{equation}
where $\mathcal{L}$ denotes the loss and $\eta$ denotes the learning rate. After fine-tuning, SparseLoRA first prunes $\Delta \W$ with binary masks and then incorporate it with the pruned weights $\W$: 
$\boldsymbol{W} \leftarrow \hat{\boldsymbol{W}} = \boldsymbol{W} + BA \odot \m$.  
The adaptation of SparseLoRA finetuning ensures the sparsity of incremental weights, thus preserving the sparse pattern after merging. Other than the vision model and language model, VLMs also involve small learnable interfaces (e.g., QFormer \cite{blip-2, instructblip}) that align vision models and language models. 
Because of this, we also insert LoRA into the QFormer, which enhances cross-modality adaptation with minimal additional computational overhead.

\subsection{Finetuning Objectives}

To recover the performance of pruned VLMs, we introduce two finetuning objectives. Firstly, acknowledging the performance gap, we continue to finetune VLMs on the pretraining tasks by minimizing loss $\mathcal{L}_\text{task}$ to restore task-specific performance. On the other hand, we propose distilling knowledge \cite{hinton2015distilling, Gou_2021, stanton2021does} from the original models to the pruned models by constraining the KL divergence between their outputs. The distillation loss $\mathcal{L}_\text{distill}$ is formulated as follows:
\begin{equation}
    \mathcal{L}_\text{distill} = 
    D_\text{KL}
    \bigg(
    \text{logits}\left(\hat\W\right) ~\|~ \text{logits}\left( \W_\text{0} \right)
    \bigg), 
    \label{eq:distill}
\end{equation}
where $D_\text{KL}$ represents the KL-divergence distance and logits$(\W_0)$ denotes the output logits of the model with weights $\W_0$. Based on the original model with weights $\W_0$, both $\text{logits}\left(\hat\W\right)$ and $\text{logits}\left( \W_\text{0} \right)$ can be obtained by forwarding with $\W_0$ and $(\W_0 + BA)\odot \m$ separately. 
This avoids hosting additional weights during training. The overall optimization objective of SparseLoRA is: 
\begin{equation}
\mathcal{L} = \lambda \mathcal{L}_{\text{task}} + (1 - \lambda)\mathcal{L}_{\text{distill}}, 
\end{equation}
where $\lambda$ is a scalar weight. 
The procedure of VLM pruning and SparseLoRA is shown in Figure \ref{fig:overview}.


\begin{table*}
    \centering
    \caption{
    \textbf{Comparison of Full Model, pruned models, and retrained pruned models on the zero-shot performance with BLIP-2 \cite{blip-2} at 50\% sparsity.} Metrics include accuracy for visual question answering, CIDEr and SPICE for image captioning, and TR@1 (text recall) and IR@1 (image recall) for image retrieval. Results are averaged over 5 runs, with the best-performing results marked in \textbf{bold} (Full Model not included). 
    }
    \vspace{-5pt}
    \label{tab:main result}
    \resizebox{0.98\textwidth}{!}{
    \begin{tabular}{lcccccccccc}
    \toprule

    \multirowcell{3}{Method} & \multirowcell{3}{Sparsity} &
    \multirowcell{3}{Param.} 
    & \multicolumn{3}{c}{Visual Question Answering}  & \multicolumn{2}{c}{Image Captioning} & \multicolumn{2}{c}{Image retrieval} & \multirowcell{3}{Macro \\ Avg.} \\
    \cmidrule{4-10}
    & 
    &
    & VQAv2 & OK-VQA & GQA & \multicolumn{2}{c}{NoCaps} & \multicolumn{2}{c}{Flickr30k} &  \\   
    &
    &
    & \multicolumn{3}{c}{Accuracy} & CIDEr & SPICE & TR@1 & IR@1 & \\   
    \midrule
    Full Model  
    & 0\% 
    & 3.9B
    & 63.1 & 41.1 & 44.1 & 105.4 & 13.8 & 96.1 & 87.5 & 64.4 \\
    \midrule
    Magnitude  
    & \multirowcell{5}
    {50\%} 
    &
    \multirowcell{5}
    {2.1B} 
    & 0.0 & 0.0 & 0.0 & 0.0 & 0.0 & 0.2 & 0.1 & 0.0 \\
    Gradient 
    &
    &
    & 55.1 & 35.7 & 39.8 & 92.3 & 11.6 & 91.4 & 81.6 & 58.2 \\
    SparseGPT 
    &  
    &  
    & 56.1 & 35.5 & 40.6 & 98.7 & 13.3 & {95.8} & {86.2} & 60.9 \\
    Wanda 
    & 
    &  
    & 57.7 & 35.4 & 41.9 & 100.1 & 13.4 & 95.2 & 84.5 & 61.2 \\
    ECoFLaP
    &  
    & 
    & 57.5 & 36.2 & 42.1 & 99.0 & 12.5 & 95.7 & 85.8 & 61.3 \\
    \midrule
    \gr
    Wanda + DS{\footnotesize \faBan}T 
    & 
    & 
    & 57.3 & 35.5 & 42.5 & 100.9 & 13.3 & 95.3 & 85.4 & 61.5 \\
    \gr
    Wanda + SparseLoRA 
    & \multirow{-2}{*}{$50\%$} 
    & \multirow{-2}{*}{$2.1$B} 
    & \bf 61.2 & \bf 39.5 & \bf 43.5 & \bf 106.6 & \bf 14.1 & \bf 96.0 & \bf 87.2 & \bf 64.0 \\
    \bottomrule
    \end{tabular}
    }
\vspace{-10pt}
\end{table*}

\section{Experimental Setup}

\textbf{Architectures. }
We use multiple multi-modal architectures for experiments including BLIP-2 \cite{blip-2} and InstructBLIP \cite{instructblip}, which composes of pretrained EVA-ViT (ViT-g/14 from EVA-CLIP) \cite{Sun2023EVACLIPIT} and pretrained large language models (i.e., FlanT5 \cite{Chung2022ScalingIL} and Vicuna \cite{vicuna2023}). 

\textbf{Evaluation Datasets and Metrics. }
We evaluate the zero-shot ability of BLIP-2 and InstructBLIP on various datasets after pruning. We use VQAv2 \cite{Goyal2016MakingTV}, OK-VQA \cite{Marino2019OKVQAAV}, and GQA \cite{Hudson2019GQAAN} for visual question answering, NoCaps \cite{Agrawal2019nocapsNO} for image captioning, and Flickr30k \cite{Plummer2015Flickr30kEC} for image-text retrieval. We use CIDEr and SPICE to evaluate image captioning tasks and use TR@1 (top-1 text recall) and IR@1 (top-1 image recall) for image retrieval tasks.

\textbf{Calibration and Training Datasets.}
Following \cite{ECoFLaP, liu2023visual}, our approach leverages a small subset of CC3M \cite{cc3m} for calibration and training data. The number of training samples ranges from 1k to 10k, while the number of calibration samples is 128, which has been shown to be sufficient for pruning \cite{wanda, sparsegpt, dsnot, ECoFLaP}.

\textbf{Finetuning Details} We use Adam~\cite{kingma2014adam} as the optimizer with $\beta_1$, $\beta_2$ = 0.9, 0.999. For regularization, we set the $\lambda$ as 0.1 and grid-search the learning rate from \{1e-5, 2e-5, 5e-5, 1e-4, 2e-4\}, where we warm up the learning rate in the first 10\% steps (of the total training steps). For different model scales, we select a batch size from \{16, 32, 64\}, and finetune 1 epoch, which is enough for convergence. We perform a grid search for the rank of SparseLoRA, considering values from \{4, 8, 16, 32\}. By trial and error, we found that a rank of 4 suffices for the QFormer and the vision model, while a rank of 8 optimally suits the language model. 

\textbf{Baselines. }
We consider several pruning techniques, including Global Magnitude Pruning, Gradient-based Pruning, SparseGPT \cite{sparsegpt}, and Wanda \cite{wanda}. Global Magnitude Pruning prunes are based on weight magnitude, while Gradient-based Pruning prunes use the product of first-order gradient and weight magnitude \cite{ECoFLaP}. SparseGPT is a layer-wise Hessian-based method, and Wanda utilizes weight magnitude and input activation norm for layer-wise pruning. Additionally, we compare against ECoFLaP \cite{ECoFLaP}, which adopts a zero-order gradient-based layer-wise sparsity for vision-language models. We also compare SparseLoRA against DS{\footnotesize \faBan}T  \cite{dsnot} that updates the masks after pruning.

\section{Results}

\begin{table*}
    \centering
    \caption{\textbf{Performance comparison at different sparse patterns} (i.e., unstructured 50\%, 2:4 and 4:8). using InstructBLIP \cite{instructblip} as the backbone. The shown results are the averaged score for 5 runs and the absolute performance gain is denoted as ${\Uparrow(\cdot)}$.
    }
    \label{tab:structured}
    \resizebox{0.96\textwidth}{!}{
    \begin{tabular}{llccccccc}
    \toprule
    &
    \multirowcell{3}{Method} & \multirowcell{3}{Sparsity} 
    & \multicolumn{3}{c}{Visual Question Answering}  & \multicolumn{2}{c}{Image Captioning} 
    & \multirowcell{3}
    {Macro \\ {Avg.}} 
    \\
    \cmidrule{4-8}
    &
    &
    & VQAv2 & OK-VQA & GQA & \multicolumn{2}{c}{NoCaps} & 
    \\   
    &
    &
    & \multicolumn{3}{c}{Accuracy} & CIDEr & SPICE & 
    \\   
    \midrule
    &
    Full Model  
    & 0\%  
    & 73.5 & 52.6 & 48.4 & 121.4 & 15.6  
    & \underline{62.3} \\
    \cmidrule{2-9}
    &
    Wanda  
    & \multirowcell{3}{50\%} 
    & 69.1 & 45.4 & 45.7 & 108.7 & 14.2 & \underline{56.6}
    \\
    &    w/DS{\footnotesize \faBan}T 
    &
    & 68.6 & 45.5 & 45.6 & 107.0 & 14.2 & \underline{56.2}  \\
    \gr
    \cellcolor{white}
    \multirow{-0}*{\shortstack{InstructBLIP \\ $\text{FlanT5}_\text{XL}$}}
    &
    w/SparseLoRA & 
    & ~$\bf 71.0^{\Uparrow +1.9}$ 
    & ~$\bf 48.4^{\Uparrow +3.0}$ 
    & ~$\bf 46.7^{\Uparrow +1.0}$ 
    & ~$\bf 118.4^{\Uparrow +9.7}$ 
    & ~$\bf 15.4^{\Uparrow +1.2}$ 
    & ~$\bf \underline{60.0}^{\Uparrow +3.4}$ 
    \\
    \cmidrule{2-9}
    &
    Wanda 
    & \multirowcell{3}{2:4} 
    & 61.2 & 33.9 & 42.1 & 82.5 & 11.9 & \underline{46.3}
    \\
    &    w/DS{\footnotesize \faBan}T 
    &  
    & 63.5 & 35.8 & 42.8 & 96.1 & 13.0 & \underline{50.2} \\
    \gr
    \cellcolor{white}
    &
    w/SparseLoRA
    &  
    & ~$\bf 67.4^{\Uparrow +6.2}$ 
    & ~$\bf 43.1^{\Uparrow +9.2}$ 
    & ~$\bf 43.8^{\Uparrow +1.7}$ 
    & ~$\bf 114.7^{\Uparrow +32.2}$ 
    & ~$\bf 14.9^{\Uparrow +3.0}$ 
    & ~$\bf \underline{56.8}^{\Uparrow +10.5}$
    \\
    \cmidrule{2-9}
    &
    Wanda
    & \multirowcell{3}{4:8} 
    & 66.0 & 39.8 & 45.1 & 97.1 & 13.1 
    & \underline{52.2} \\
    & w/DS{\footnotesize \faBan}T 
    &  
    & 67.3 & 41.4 & 46.3 & 105.7 & 13.9 & \underline{54.9} \\
    \gr        
    \cellcolor{white}
    & w/SparseLoRA
    &  
    & ~$\bf 69.4^{\Uparrow +3.4}$ 
    & ~$\bf 45.0^{\Uparrow +5.2}$ 
    & ~$\bf 46.9^{\Uparrow +1.8}$ 
    & ~$\bf 116.1^{\Uparrow +19.0}$ 
    & ~$\bf 15.1^{\Uparrow +2.0}$ 
    & $\bf \underline{58.5}^{\Uparrow +6.2}$ \\
    \cmidrule{1-9}
    &
    Full Model
    & \multirowcell{1}{0\%} 
    & 76.7 & 58.8 & 49.1 & 123.9 & 15.9 
    & \underline{64.9} \\
    \cmidrule{2-9}
    &
    Wanda
    & \multirowcell{3}{50\%} 
    & 67.7 & 47.8 & 44.9 & 109.7 & 14.6 
    & \underline{56.9} \\
    & 
    w/DS{\footnotesize \faBan}T 
    &  
    & 67.5 & 47.6 & 44.8 & 109.3 & 14.6 & \underline{56.8} \\
    \gr
    \cellcolor{white}
    \multirow{-0}*{\shortstack{InstructBLIP \\ Vicuna-7B}}
    &
    w/SparseLoRA
    &
    & ~$\bf 72.2^{\Uparrow +4.5}$ 
    & ~$\bf 52.0^{\Uparrow +4.2}$ 
    & ~$\bf 48.3^{\Uparrow +3.4}$ 
    & ~$\bf 118.2^{\Uparrow +8.5}$ 
    & ~$\bf 15.1^{\Uparrow +0.5}$ 
    & $\bf \underline{61.2}^{\Uparrow +4.3}$ \\
    \cmidrule{2-9}
    &
    Wanda
    & \multirowcell{3}{2:4} 
    & 58.7 & 32.1 & 39.0 & 68.8 & 12.9 
    & \underline{42.3} \\
    & w/DS{\footnotesize \faBan}T 
    &  
    & 60.2 & 32.3 & 41.4 & 66.9 & 12.6 & \underline{42.7} \\
    \gr
    \cellcolor{white}
    & w/SparseLoRA
    &  
    & ~$\bf 66.2^{\Uparrow +7.5}$ 
    & ~$\bf 43.6^{\Uparrow +11.5}$ 
    & ~$\bf 44.5^{\Uparrow +5.5}$ 
    & ~$\bf 112.2^{\Uparrow +43.4}$ 
    & ~$\bf 14.6^{\Uparrow +1.7}$ 
    & $\bf \underline{56.2}^{\Uparrow +13.9}$ \\
    \cmidrule{2-9}
    &
    Wanda
    & \multirowcell{3}{4:8} 
    & 61.4 & 39.5 & 42.4 & 95.5 & 13.6 
    & \underline{50.5} \\
    & w/DS{\footnotesize \faBan}T 
    &  
    & 63.3 & 39.6 & 44.6 & 101.1 & 13.9 & \underline{52.5} \\
    \gr
    \cellcolor{white}
    &
    w/SparseLoRA
    &  
    & ~$\bf 69.5^{\Uparrow +8.1}$ 
    & ~$\bf 47.4^{\Uparrow +7.9}$ 
    & ~$\bf 45.8^{\Uparrow +3.4}$ 
    & ~$\bf 115.1^{\Uparrow +19.6}$ 
    & ~$\bf 14.9^{\Uparrow +1.3}$ 
    & $\bf \underline{58.5}^{\Uparrow +8.0}$ \\
    \bottomrule
    \end{tabular}
    }
\vspace{-2pt}
\end{table*}

\subsection{Main Experimental Results}
\textbf{Unstructured Sparsity. }
In Table \ref{tab:main result}, We compare the zero-shot performance on various datasets using BLIP-2 pruned by different pruning techniques at unstructured 50\% sparsity ratios. Among all pruning methods, while Wanda and ECoFLaP achieve the best performance, Wanda does not require multiple forward passes and is much more time-efficient. On the other hand, considering, EcoFLaP does not apply for N:M sparsity, we use Wanda as the default pruning method.

Compared to DS{\footnotesize \faBan}T that focuses on reconstruction errors, SparseLoRA also considers task-specific performance and knowledge distillation from original full models, consistently outperforming the baselines on all tasks. Notably, the average performance of SparseLoRA is comparable to that of the full model. 

\begin{table}
    \centering
    \caption{\textbf{Performance comparison of pruning single modality} on InstructBLIP-Vicuna-7B.}
    \label{tab:single}
    \vspace{-0.1in}
    \resizebox{\linewidth}{!}{
    \begin{tabular}{lccccc}
    \toprule    
    \multirowcell{2}{Method} & \multirowcell{2}{Param.} & \multicolumn{2}{c}{VQA} & \multicolumn{2}{c}{NoCaps} \\
    & 
    & VQAv2 & GQA & CIDEr & SPICE \\
    \midrule
    Full Model 
    & 7.9B
    & 76.7 & 49.1 & 123.9 & 15.9  \\
    \midrule
    \multicolumn{6}{c}{\bf \textit{2:4 Sparsity}} \\
    \hdashline
    Wanda 
    & \multirowcell{3}{4.7B}
    &
    60.5 & 41.2 & 110.2 & 15.4  \\
    w/DS{\footnotesize \faBan}T 
    & 
    & 
    64.9 & 43.5 & 107.2 & 14.8  \\
    w/SparseLoRA 
    & 
    & 
    \bf 68.3 & \bf 45.4 & \bf 119.3 & \bf 15.5 \\
    \midrule
    \multicolumn{6}{c}{\bf \textit{4:8 Sparsity}} \\
    \hdashline
    Wanda 
    & \multirowcell{3}{4.7B}
    & 63.9 & 43.1 & 116.0 & 15.4 \\
    w/DS{\footnotesize \faBan}T 
    & 
    & 
    68.3 & 44.8 & 115.2 & 15.1  \\
    w/SparseLoRA
    & 
    & 
    \bf 71.4 & \bf 46.5 & \bf 121.6 & \bf 15.6 \\
    \bottomrule
    \end{tabular}
    }
    \vspace{-10pt}
\end{table}

\textbf{N:M Sparsity. } 
In addition to unstructured sparsity, we also conduct experiments on N: M sparsity \cite{zhou2021, zhang2022learning}, which can be applied to specific GPU cores and has more practical applications \cite{mishra2021accelerating}. Compared to unstructured pruning, structured pruning causes a more significant performance drop and requires more extensive restores. Under more structured patterns, SparseLoRA recovers more performance, achieving a 10.5\% improvement for 2:4 sparsity compared to 3.4\% for unstructured sparsity. After restoring, all structured pruned models maintain over 90\% of the performance of the original models, demonstrating the universality and effectiveness of SparseLoRA.

\textbf{Single Model Pruning. }
Language models typically have much larger parameter sizes compared to the vision models in vision-language models, \cite{blip-2, instructblip} (e.g., 7B for Vicuna \cite{vicuna2023} vs. 1.3B for EVA-ViT in parameters \cite{Sun2023EVACLIPIT}). As a result, the efficiency bottleneck primarily stems from the language model component. This prompted us to investigate the impact of solely pruning language models in VLMs, with experimental results presented in Table \ref{tab:single}.
With additional parameters in the vision model component, SparseLoRA restores InstructBLIP with significant improvement (e.g., from 69.0 to 71.6 on VQAv2), achieving performance comparable to the Full Model. Therefore, pruning language models only is an effective way to maintain performance and efficiency. 

\begin{table}
    \centering
    \caption{
    Comparison between LoRA and SparseLoRA. 
    }
    \label{tab:sparselora}
    \vspace{-0.1in}
    \resizebox{\linewidth}{!}{
    \begin{tabular}{lcccc}
    \toprule    
    Method & Sparsity & VQAv2 & OK-VQA & GQA \\
    \midrule
    Full Model 
    & 0\% & 76.7 & 58.8 & 49.1 \\
    \hdashline
    LoRA 
    &  & \bf 74.1 & 52.9 & 48.2 \\
    \gr
    SparseLoRA 
    & \multirow{-2}*{50\%} & 74.0 & \bf 53.3 & \bf 48.6 \\
    \hdashline
    LoRA 
    &  & 67.8 & 43.7 & 44.9 \\
    \gr
    SparseLoRA 
    & \multirow{-2}*{2:4} & \bf 68.3 & \bf 44.6 & \bf 45.4 \\
    \hdashline
    LoRA 
    &  & 70.2 & 48.3 & 45.9 \\
    \gr
    SparseLoRA 
    & \multirow{-2}*{4:8} & \bf 71.4 & \bf 49.2 & \bf 46.5 \\
    \bottomrule
    \end{tabular}}
\end{table}

\subsection{Detailed Analysis}

\begin{table*}[h]
    \centering
    \caption{\textbf{Performance of SparseLoRA applied on pruning scenarios}, where ``Vision + Language'' denotes pruning both vision models and language models, and ``Language'' denotes pruning language models only. $V$, $L$, $Q$ represent the models for SparseLoRA. }
    \label{tab:single-modality}
    \vspace{-0.1in}
    \resizebox{\linewidth}{!}{
    \begin{tabular}{lc
cccccccc}
    \toprule    
    \multirow{2}{*}{Method} & ~\multirow{2}{*}{Modality}~ & \multicolumn{4}{c}{Vision + Language} & \multicolumn{4}{c}{Language} \\ 
    \cmidrule(lr){3-6}
    \cmidrule(lr){7-10}
    &  & ~VQAv2~ & ~OK-VQA~ & ~GQA~ & ~\underline{Avg. }~  &
    ~VQAv2~ & ~OK-VQA~ & ~GQA~ & ~\underline{Avg. }~ \\
    \midrule
    Wanda 
    & 
    --
    & 61.4 & 39.5 & 42.4 & \underline{47.8} 
    & 63.9 & 44.5 & 43.1 & \underline{50.5}
    \\
    \midrule
    \multirowcell{5}{w/SparseLoRA} 
    & $V$
    & 64.3 & 42.6 & 45.8 & \underline{51.0}
    & 66.0 & 44.7 & 43.3 & \underline{51.3}
    \\
    & $L$
    & 66.4 & 46.7 & 44.3 & \underline{52.5}
    & \bf 70.8 & \bf 46.5 & \bf 49.5 & \underline{\bf 55.6}
    \\
    & $Q$
    & 62.5 & 40.2 & 43.3 & \underline{48.7}
    & 64.3 & 44.5 & 44.1 & \underline{51.0}
    \\
    & $V+L$
    & \bf 69.5 & \bf 47.4 & \bf 45.8 & \underline{\bf 54.2}
    & 70.6 & 46.3 & 49.3 & \underline{55.4}
    \\
    &  $V+L+Q$
        & 69.0 & 46.9 & 45.4 & \underline{53.8}
    & 70.2 & 45.9 & 48.1 & \underline{54.7}
    \\
    \bottomrule
    \end{tabular}}
\label{tab:cross-modality}
\end{table*}

To evaluate cross-modality adaptation, we integrate SparseLoRA into various models. Specifically, we denote the QFormer, vision model, and language model as "$Q$", "$V$", and "$L$" respectively. Different configurations are represented using combinations of these notations (e.g., "$QLV$" and "$LV$"). As shown in Table \ref{tab:cross-modality}, for cross-modality pruning (i.e., Vision + Language), finetuning within a single model contributes to performance restoration, while finetuning models across two modalities further enhances performance. In cases of single modality pruning, finetuning the pruned model alone is sufficient for restoration. Notably, joint finetuning with the QFormer does not yield performance gains beyond finetuning the pruned models. 


\begin{table}
    \centering
    \caption{Ablation studies on different finetuning objectives.}
    \label{tab:kd}
    \vspace{-0.1in}
    \resizebox{\linewidth}{!}{
    \begin{tabular}{lcccc}
    \toprule    
    \multirowcell{2}{Method} & \multicolumn{2}{c}{Flickr30k} & \multicolumn{2}{c}{NoCaps} \\
    & TR@1 & IR@1 & CIDEr & SPICE \\
    \midrule
    Full Model & 96.1 & 87.5 & 105.4 & 13.8  \\
    \hdashline
    $\mathcal{L}_\text{task}$ & 95.3 & 86.2 & 106.1 & 14.0 \\
    $\mathcal{L}_\text{distill}$ & 95.4 & 86.6 & 102.2 & 13.4 \\
    \gr
    $\mathcal{L}_\text{task} \& \mathcal{L}_\text{distill}$ & \bf 96.0 & \bf 87.2 & \bf 106.6 & \bf 14.1 \\
    \bottomrule
    \end{tabular}}
\end{table}

\textbf{SparseLoRA Finetuning Achieves Comparable Performance with LoRA. } 
LoRA weights cannot be merged with pruned weights, as this would disrupt the sparse pattern. Consequently, the presence of remaining LoRA modules leads to latency and slows down inference significantly \cite{dery2024everybody, rücklé2021adapterdrop}. To address this issue, SparseLoRA aims to resolve the unmerged weights of LoRA and eliminate the latency caused by LoRA modules. Table \ref{tab:sparselora} compares the performance of LoRA with SparseLoRA for VLMs with sparse language models. Remarkably, SparseLoRA finetuning achieves improved performance with fewer trainable parameters, consistent with findings from \cite{sparseadapter}.

\textbf{The Effectiveness of Finetuning Objectives. }
We further investigate the impact of the proposed finetuning objectives on BLIP-2-$\text{FlanT5}_\text{XL}$. In Table \ref{tab:kd}, we consider three finetuning objectives: $\mathcal{L}_\text{task}$, $\mathcal{L}_\text{distill}$, and $\mathcal{L}_\text{task} \& \mathcal{L}_\text{distill}$. $\mathcal{L}_\text{task}$ guides the task-specific performance while $\mathcal{L}_\text{distill}$ guides knowledge transferring from the original full model to the pruned dense model. either minimizing $L_\text{task}$ or $\mathcal{L}_\text{distill}$ improves the performance. In addition, jointly minimizing $\mathcal{L}_\text{distill}$ and $\mathcal{L}_\text{task}$ helps the pruned models further recover performance.

\begin{figure}[h]
  \caption{Ablation study on sparsity ratios.}
  \hspace{-6pt}
\includegraphics[width=0.45\textwidth]{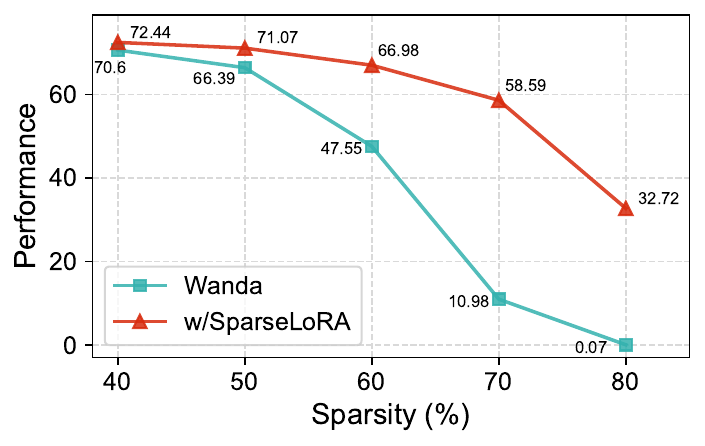}
  \label{fig:sparsity}
  \vspace{-5pt}
\end{figure}

\textbf{Ablation Study on Sparsity.} To assess the effectiveness of SparseLoRA across a broader range of sparsity ratios, we experimented on InstructBLIP-Vicuna-7B with unstructured sparsity ratios ranging from 40\% to 80\%. As the sparse ratio $s$ exceeded 50\%, the performance of pruned models began to deteriorate, eventually collapsing when $s \geq 70\%$, highlighting the necessity of restoring. In such scenarios, SparseLoRA significantly improved performance, particularly for higher sparsity ratios, achieving a recovery of 47.6\% of scores at $s=70\%$ and 32.7\% at $s=80\%$. 

\begin{figure}[h]
  \caption{Impact of finetuning samples.}
  \hspace{2pt}
  \includegraphics[width=0.45\textwidth]{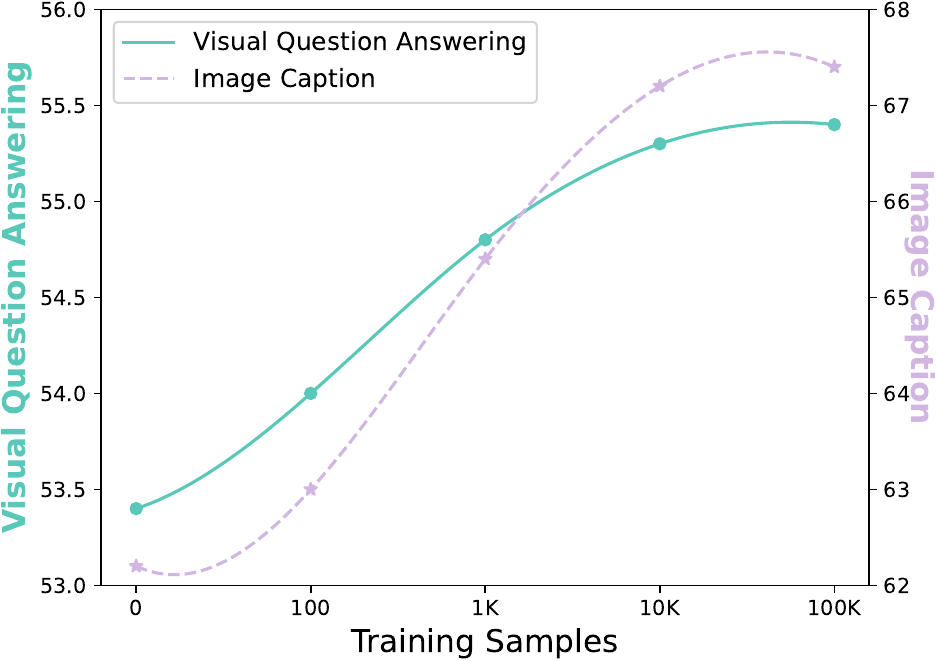}
  \label{fig:scaling_law}
  \vspace{-5pt}
\end{figure}

\textbf{Ablation Study on Calibration Datasets. }
SparseLoRA utilizes calibration datasets for retraining. We conducted experiments to explore the impact of the number of training samples. Specifically, we randomly sampled $k$ ($k$ = 0, 100, 1k, 10k, 100k) training data points from CC3M \cite{cc3m} to finetune $\text{InstructBLIP-FlanT5}_\text{XL}$ with 50\% sparsity and report the average performance of visual question answering and image caption. As shown in Figure \ref{fig:scaling_law}, we found that finetuning pruned VLMs with few-shot samples (i.e., 100) can improve performance by a substantial margin. Further finetuning with 10k training data points resulted in a significant boost in cross-modality ability. This suggests that a small amount of data is sufficient to restoration the pruned vision-language models, leveraging the knowledge and capabilities acquired during pretraining \cite{lima}. When $k \geq$ 10k, the model's capability continues to improve with more training data and gradually becomes saturated.


\section{Conclusion}
In this paper, motivated by the challenges associated with deploying VLMs in real-world applications, we investigate the potential of pruning VLMs. Specifically, recognizing that VLMs encompass models from different modalities, we conduct empirical studies to explore the distribution of sparsity ratios across these models and how sparsity impacts performance, thereby highlighting the necessity of restoring pruned VLMs. Subsequently, we introduce MAF, which addresses this challenge by restoring pruned VLMs through cross-modality adaptation and SparseLoRA finetuning. Extensive experiments validate the effectiveness of MAF, providing valuable insights for future research on VLM sparsity. 

\section{Limitations}
Despite our progress, limitations remain in our work. Although our proposed methods are universal for all VLM models, we have primarily focused on BLIP family models and selected tasks . We believe our methods can be easily extended to a broader range of models and tasks. On the other hand, given there may be potentially high-quality dataset for restoring pruned models, we believe the incorporation of such datasets would further promotes our proposed methods


\bibliography{custom}




\end{document}